\documentclass[]{sjtu_epic_blue}  
\usepackage{enumitem}
\usepackage[toc,page,header]{appendix}
\usepackage[utf8]{inputenc}
\usepackage[T1]{fontenc}
\usepackage{hyperref}
\usepackage{url}
\usepackage{booktabs}
\usepackage{amsfonts}
\usepackage{nicefrac}
\usepackage{makecell}
\usepackage{microtype}
\usepackage{graphicx}
\usepackage{multirow}
\usepackage{pifont}
\usepackage{amsmath,amssymb,amsthm}
\usepackage{etoolbox}
\usepackage{lipsum}  
\usepackage{minitoc}
\usepackage{tablefootnote}
\usepackage{threeparttable}
\usepackage{wrapfig}
\usepackage{appendix}
\usepackage{multirow}
\usepackage{ulem}
\useunder{\uline}{\ul}{}
\usepackage{colortbl}
\usepackage{xspace}
\usepackage{longtable}
\usepackage{color}
\usepackage{xcolor}
\usepackage[most]{tcolorbox}
\usepackage[utf8]{inputenc}
\DeclareUnicodeCharacter{2660}{\spadesuit}
\usepackage[most]{tcolorbox}
\usepackage{listings}
\usepackage{fontawesome5}  
\usepackage{xcolor}       
\usepackage{caption}
\usepackage{listings}
\definecolor{boxback}{gray}{0.95}
\definecolor{promptback}{gray}{0.98}
\definecolor{outputback}{RGB}{230, 245, 255}
\definecolor{titleblue}{RGB}{0, 110, 180}

\newtcolorbox{casestep}[2][]{
    enhanced,
    breakable,
    colback=boxback,
    colframe=gray!70,
    coltitle=white,
    colbacktitle=titleblue,
    fonttitle=\bfseries,
    title=#2,
    attach boxed title to top left={yshift=-2mm, xshift=4mm},
    boxed title style={sharp corners, boxrule=0pt},
    before skip=15pt,
    #1
}
\usepackage{amsmath}
\allowdisplaybreaks  
\lstdefinestyle{promptstyle}{
    backgroundcolor=\color{promptback},
    basicstyle=\ttfamily\small,
    breaklines=true,
    frame=tb,
    framerule=0pt,
    rulecolor=\color{gray!40},
    xleftmargin=1em,
    xrightmargin=1em,
      xleftmargin=0pt,            
      xrightmargin=0pt,           
      framesep=3pt,               
      rulesep=2pt,
}

\newtcolorbox{modeloutput}[2][]{
    enhanced,
    breakable,
    colback=outputback,
    colframe=titleblue!50,
    boxrule=1pt,
    fonttitle=\bfseries\small,
    coltitle=titleblue,
    title=#2,
    #1
}
\newtcolorbox{modeloutput_continue}[1][]{
    enhanced,
    breakable,
    colback=outputback,
    colframe=titleblue!50,
    boxrule=1pt,
    #1
}

\title{AI for Service:  Proactive Assistance with AI Glasses}

\makeatletter
\gdef\authorlist{%
  \begin{center}
    \authorfont\sffamily
    Zichen Wen$^{1*}$,  
    Yiyu Wang$^{1*}$,
    Chenfei Liao$^{2,1,3*}$,
    Boxue Yang$^{1*}$,
    Junxian Li$^{1*}$,
    Weifeng Liu$^{4,1*}$ \\
    Haocong He$^{1}$, 
    Bolong Feng$^{1}$, 
    Xuyang Liu$^{1}$,
    Yuanhuiyi Lyu$^{2,3}$,
    Xu Zheng$^{2,3}$ \\
    Xuming Hu$^{2,3}$,
    Linfeng Zhang$^{1\dag}$
  \end{center}
}
\makeatother

\makeatletter
\gdef\affiliationlist{%
  \begin{center}
    \affiliationfont\sffamily
    $^{1}$EPIC Lab, Shanghai Jiao Tong University\\
    $^{2}$The Hong Kong University of Science and Technology (Guangzhou)\\
    $^{3}$The Hong Kong University of Science and Technology\\
    $^{4}$Peking University
  \end{center}
}
\makeatother

\abstract{
    In an era where AI is evolving from a passive tool into an active and adaptive companion, we introduce \textit{\textbf{AI for Service}} (AI4Service), a new paradigm that enables proactive and real-time assistance in daily life. Existing AI services remain largely reactive, responding only to explicit user commands. We argue that a truly intelligent and helpful assistant should be capable of anticipating user needs and taking actions proactively when appropriate. 
    To realize this vision, we propose \textit{\textbf{Alpha-Service}}, a unified framework that addresses two fundamental challenges: \textit{Know When} to intervene by detecting service opportunities from egocentric video streams, and \textit{Know How} to provide both generalized and personalized services. Inspired by the von Neumann computer architecture and based on AI glassess, Alpha-Service consists of five key components: an \textit{Input Unit} for perception, a \textit{Central Processing Unit} for task scheduling, an \textit{Arithmetic Logic Unit} for tool utilization, a \textit{Memory Unit} for long-term personalization, and an \textit{Output Unit} for natural human interaction. 
    As an initial exploration, we implement Alpha-Service through a multi-agent system deployed on AI glasses. Case studies, including a real-time Blackjack advisor, a museum tour guide, and a shopping fit assistant, demonstrate its ability to seamlessly perceive the environment, infer user intent, and provide timely and useful assistance without explicit prompts. 
}

\date{\today}

\begin{document}
\maketitle

{
\renewcommand{\thefootnote}{\fnsymbol{footnote}}
  \footnotetext[1]{Equal Contribution and Core Contributors.} 
  \footnotetext[2]{Corresponding author: zhanglinfeng@sjtu.edu.cn}
}

\begin{figure}[h]
    \centering
\includegraphics[width=\linewidth]{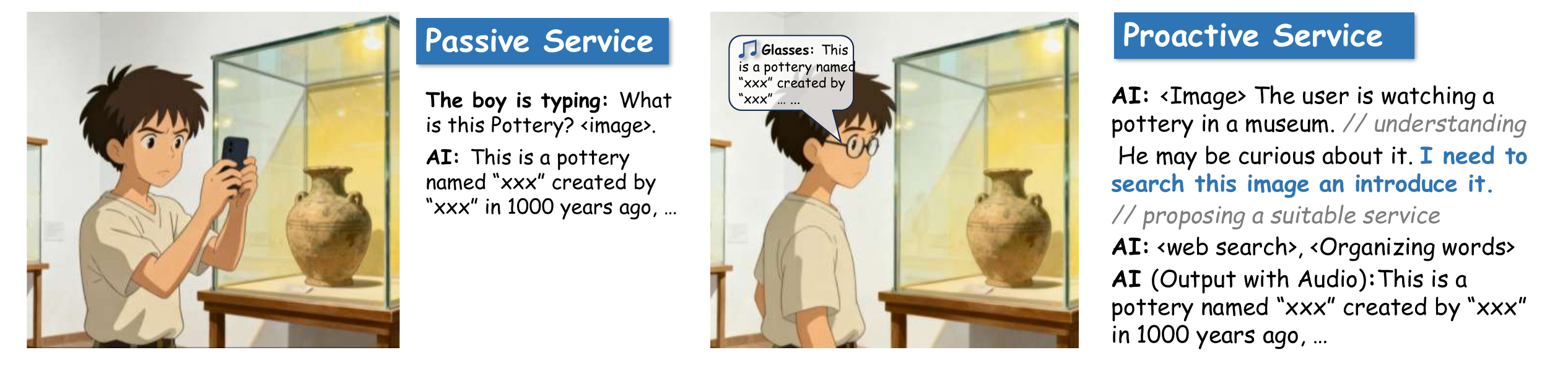}
    \caption{Comparison between passive service and proactive service. Passive service indicates that AI provides service only when it is asked for, while proactive service indicates that the AI keeps observing the environment, guesses the user's thoughts, consider the user's needs, and provide services without the user having to ask.} 
    \label{fig:head_fig}
\end{figure}

\section{Introduction}
\label{sec:intro}

Artificial intelligence has long been envisioned to enhance the quality of human life. In early research, through technologies such as image recognition~\citep{shih2010image} and natural language understanding~\citep{allen1995natural}, AI has already provided effective services for specific tasks like autonomous driving~\citep{yurtsever2020survey,xiong2025prune2drive} and machine translation~\citep{poibeau2017machine,li2025tactic}. In recent years, breakthroughs in large language models~\citep{achiam2023gpt,bai2023qwen,team2025kimi,grattafiori2024llama} and multimodal large models~\citep{team2025kimi-vl,bai2025qwen2-vl,wen2025efficient} have significantly increased AI's potential for service provision in general scenarios. Concurrently, the proliferation of hardware devices such as AI speakers~\citep{kim2022human}, headphones, and glasses~\citep{waisberg2024meta} has made real-time interaction between AI and humans feasible. Against the backdrop of increasingly mature model capabilities and hardware foundations, AI-powered human services are undergoing a profound transformation.

However, most existing service models remain predominantly passive, requiring users to issue explicit commands before AI addresses problems in a predefined manner. This paradigm limits the deeper application of AI in the service domain and hinders its seamless integration into daily life~\citep{weiser1991computer,schilit1994context,dey2001context,horvitz1999mixed,pejovic2015anticipatory}. To address this, this paper proposes the concept of ``AI4Service'', aiming to leverage AI technology to serve all aspects of human life. We posit that realizing this vision requires focusing on three core characteristics:
\begin{itemize}[leftmargin=*,labelsep=1em]
    \item 
    \textbf{Generalization:} AI should function as a general-purpose assistant~\citep{li2024multimodal}, capable of handling diverse challenges in life rather than being confined to specific tasks. Specifically, the system should not require pre-definition or specialized training for every task, but instead rely on the inherent strong generalization capabilities of large models and the self-evolution properties of agent systems~\citep{tao2024survey}. Recent research indicates that agents can already autonomously plan and execute actions towards given goals, providing a feasible pathway towards generalizable service.

    \item     
    \textbf{Proactivity:} 
    AI should transition from passively receiving instructions to actively discovering and delivering services.
    This requires the system to continuously observe the environment, understand the user’s behavior and intent, record relevant contextual information, and proactively infer potential user needs.
    The core idea is to shift the point of service intervention from ``after the user asks'' to ``when the user’s need arises''.

    \item 
    \textbf{Customization:} Given individual differences in values, lifestyles, and privacy preferences, AI services must be deeply adaptable to individual needs. By incorporating long-term memory mechanisms, such as an Agent Memory architecture~\citep{wang2024agent}, the system can continuously learn user habits and preferences, dynamically adjusting service strategies and content to achieve a highly personalized service experience.

\end{itemize}

With the concurrent advancement of model capabilities~\citep{feng2024far}, agent technology~\citep{jennings1998roadmap,team2025kimi-rl}, and hardware platforms exemplified by AI glasses, we believe the current period presents a critical opportunity for realizing AI4Service. This paper proposes a foundational framework named ``Alpha Service'' to address this challenge. Inspired by the von Neumann computer architecture~\citep{goldstine1993computer}, this framework comprises the following five core components:

\begin{itemize}[leftmargin=*,labelsep=1em]
    \item 
\textbf{Input Unit:} Equipped with a multimodal large model capable of understanding first-person video streams, responsible for continuously perceiving the physical world and user state.
    \item 
\textbf{Central Processing Unit (CPU):} Serves as the system's control center, responsible for task parsing and scheduling. For instance, it determines the required service type based on input information and coordinates other modules to complete the task.
    \item 
\textbf{Memory Unit:} Dedicated to the persistent storage of user historical interactions and preference information, supporting efficient data writing and retrieval.
    \item 
\textbf{Arithmetic Logic Unit (ALU):} Provides various task execution tools, which can be specialized models, large models, or web search engines, responsible for executing and computing specific tasks.
    \item 
\textbf{Output Unit:} Summarizes and presents the results in user-friendly formats, such as speech or concise text. It can also choose to output nothing in some settings.
\end{itemize}
Through the coordinated operation of these components, we have successfully developed an agent system embodied in AI glasses. This system can proactively identify service opportunities and provide solutions without requiring human intervention. For example, in a game of Blackjack, the system can analyze the situation and proactively offer strategic advice to the player on whether to request another card. The detailed design principles and experimental validation of the system will be elaborated in subsequent sections.


\section{Concept of AI for Service}

\subsection{Definition and Key Layers}
``AI4Service'' is an emerging paradigm of intelligent services, the core of which lies in enabling AI systems to \textbf{proactively, timely, and personally} respond to users' needs, much like a close assistant with foresight and insight.
It transcends the traditional ``interchangeably ask-and-answer'' interaction models, aiming to anticipate service opportunities and generate corresponding service content by deeply understanding the user's current context, behavioral intentions, and long-term preferences, even before the user has explicitly expressed a need.
The objective of ``AI4Service'' is to fundamentally enhance the smoothness and satisfaction of user experiences, \textbf{achieving a transition from ``People Seek Services'' to ``AI Agents Seek Services''}.
To achieve this goal, a mature ``AI4Service'' system should possess two core layers, forming its basic architecture: 
\textbf{\ding{182} Know When:} Event Prediction and Timing. 
\textbf{\ding{183} Know How:} Generalized and Personalized Services.

\subsection{Know When: Event Prediction and Timing}
``Know When'' is the triggering mechanism and prerequisite for AI for Service. 
It requires the system to continuously perceive and analyze real-time data streams and the environment (such as video, audio, etc.), in order to accurately predict or identify characteristic timestamps that need to be provided with service.

The technical challenges at this level mainly manifest in two aspects:
\begin{itemize}[leftmargin=*,labelsep=1em]
    \item \textbf{Accurate prediction of event changes:} The system needs to detect meaningful points of state change from continuous data streams.
    For example, in a streaming video scenario, this could involve identifying when a user stops watching films. This action refers to a state change--from watching films to a new event.

    \item  \textbf{Timely classification of event types:} Once a change is detected, the system must quickly and accurately determine the type of event to match the corresponding service. 
    For instance, distinguishing whether the user's stopping is due to the new event ``attending a phone call'' or ``temporarily stepping away''—different types of new events will trigger completely distinct service responses.
\end{itemize}

The essence of ``Know When'' is to achieve the optimal timing for service, balancing between avoiding service delays that could frustrate users and preventing unnecessary frequent interruptions. This relies on high-precision temporal pattern recognition and context-aware technologies.

\subsection{Know How: Generalized and Personalized Services}
“Know How” represents 
the execution layer of AI for Service. 
Once the service timing and event type are determined, the system needs to generate concrete, useful, and user-aligned service content. 
Depending on the scope and depth of the context information relied upon, service strategies can be divided into two levels:

\begin{itemize}[leftmargin=*,labelsep=1em]
     \item \textbf{Generalized Services: }
     Generalized services are based on the immediately occurring ``event type'' and ``short-term context''. 
     They do not take the user's personal history into account, but provide standardized and universal service options for all users for a certain type of event. 
     The advantage of such services lies in their quick response and relatively low development cost, addressing the common needs of most users in specific scenarios.
     The service triggered for all users at this moment is the same.
     
     \textit{\underline{Example}: When the system detects that a user arrives at an unfamiliar outdoor location, based on the scene (short-term context) and the event type (probably ``travel''), the system would universally inform the user, ``This is Cinque Terre in Italy,'' and provide related encyclopedia links or travel guides.}

     \item \textbf{Personalized Services: }
     Personalized services take a step further by deeply integrating the user's ``long-term context'' and ``repetitive behavior patterns''. By analyzing the user's historical interaction data, long-term preferences, and habits, the system can provide unique, highly customized services, significantly enhancing user engagement.
     This service, grounded in a deep understanding of user habits, achieves a better ``anticipation of user needs''.

     \textit{\underline{Example}: Similarly, when a user arrives at Cinque Terre, in addition to providing generalized information, the system can offer personalized services based on the user's long-term context (for instance, historical search records indicating plans for a European vacation next summer, multiple previous viewings of culinary documentaries, and a habit of purchasing wine). 
     In this case, the system might proactively suggest: ``I noticed your interest in European travel and cuisine, and have curated a selection of specialty restaurants and local wine tasting routes near Cinque Terre for you.'' }
     
\end{itemize}

In conclusion, AI for Service, through the organic combination of ``Know When'' and ``Know How'', and the subsequent layering from generalized to personalized services, ultimately constructs an intelligent, seamless, user-centric next-generation service ecosystem.

\section{Architecture}
\label{sec:architecture}
\begin{figure}
    \centering
    \includegraphics[width=\linewidth]{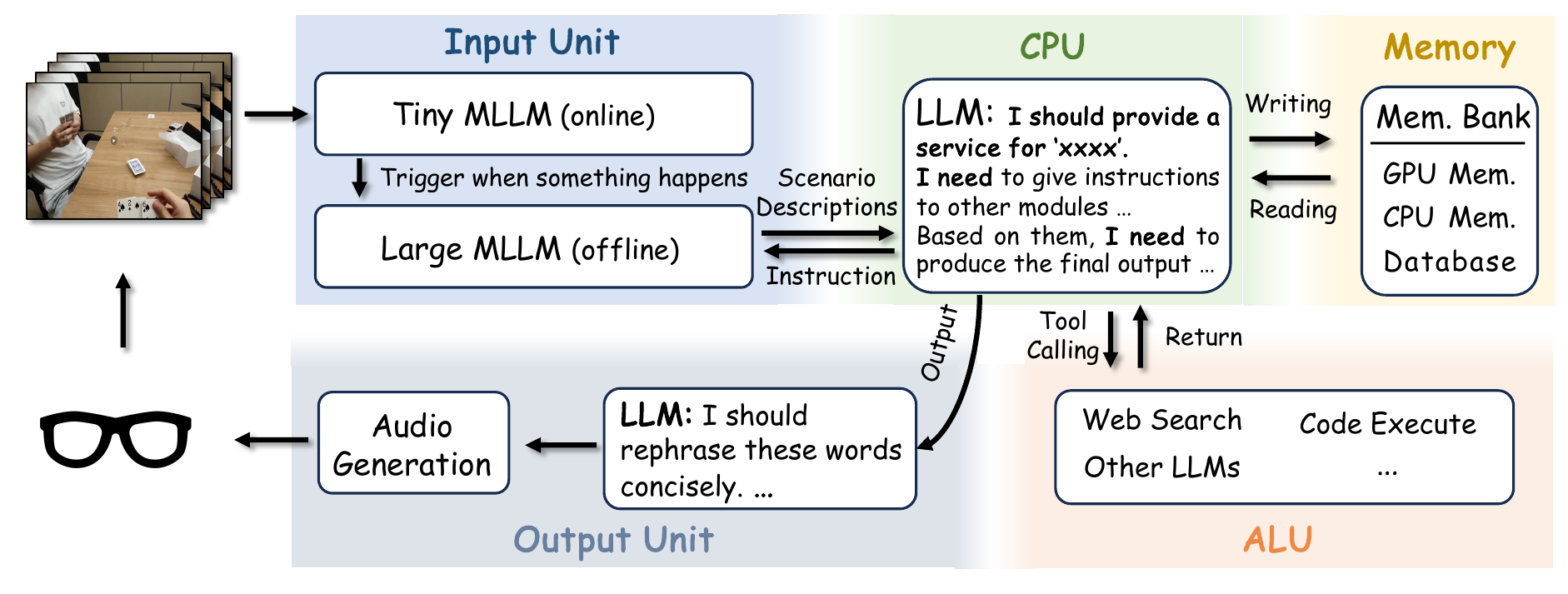}
    \caption{The architecture of Alpha-Service. }
    \label{fig:arch}
\end{figure}
\subsection{Overview: Von Neumann-Inspired Design}
Inspired by the Von Neumann paradigm, our Alpha-Service system follows a simple, modular flow: perception, dispatch, computation, memory, and delivery. Concretely, it comprises five units—Input, Central Processing (task dispatch via LLM), Arithmetic Logic (tool use), Memory (long-term context), and Output (human-friendly synthesis). The CPU orchestrates data and control among these units, enabling both reactive and proactive service assistance. Detailed designs of each unit follow in the subsequent subsections.
\subsection{Input Unit: Trigger and Streaming MLLMs}
The Input Unit serves as the agent's primary interface with the physical world, responsible for perceiving and processing real-time multi-modal data streams. At its core, this unit employs a sophisticated dual-model architecture to balance real-time responsiveness with deep scene understanding. The first component is a lightweight, continuously-running ``trigger'' model, a fine-tuned Qwen2.5-VL-3B~\citep{bai2025qwen2_5-vl}, which directly processes the video stream from the agent's first-person perspective glasses. We designed an efficient ``user command + intent'' dual-trigger mechanism, where this online model continuously analyzes incoming data for user assistance cues. Upon detecting a trigger, it sends an activation signal and preliminary scene information to the Central Processing Unit, simultaneously invoking the second component: a powerful, original Qwen2.5-VL-7B model. This larger, offline MLLM then performs a deep, fine-grained analysis of the relevant scene to provide a comprehensive understanding for decision-making. This hierarchical approach enables the agent to maintain continuous environmental perception efficiently, while leveraging powerful analytical capabilities on demand.

\subsection{Central Processing Unit: Task Orchestration and Synthesis}
The Central Processing Unit (CPU) acts as the central nervous system and reasoning core of the multi-agent system. It is responsible not only for decomposing complex user requests into executable sub-tasks but also for \textbf{collecting, integrating, and synthesizing the results} from various specialized units to formulate a final, coherent response. At the heart of the CPU is an advanced Large Language Model (LLM), fine-tuned from Qwen3-8B~\citep{yang2025qwen3}, which serves as the system's primary \textbf{Orchestrator}.

The operation of the CPU can be conceptualized in two primary phases:

\begin{enumerate}[leftmargin=*,labelsep=1em]
    \item \textbf{Decomposition and Dispatch:} Upon receiving pre-analyzed user intent and contextual data from the Input Unit, the Orchestrator LLM first evaluates the query's complexity. It then breaks the query down into a sequence of discrete, executable sub-tasks. Following this decomposition, each sub-task is routed to the most suitable specialized unit based on its requirements. This routing process includes:
    \begin{itemize}[leftmargin=*,labelsep=1em]
        \item Direct generation of a response for straightforward queries, subsequently managed by the Output Unit for human-friendly formats.
        \item Activation of a trigger model to identify the optimal timing for responses required at a designated future time step.
        \item Invocation of a streaming video LLM to produce detailed, task-specific visual descriptions when finer-grained information is necessary.
        \item Dispatch to the Arithmetic Logic Unit (ALU) for external tool invocation (e.g., web search) in cases requiring additional knowledge.
        \item Instruction to the Memory Unit for retrieval of pertinent historical interaction data.
    \end{itemize}

    \item \textbf{Synthesis and Response Generation:} After dispatching the sub-tasks, the CPU acts as a central hub to \textbf{gather the outputs} from the activated units. For instance, it may receive a detailed visual description from the video LLM, search results from the ALU, and relevant past interactions from the Memory Unit. The Orchestrator LLM then integrates these disparate pieces of information, resolves any potential conflicts, and \textbf{synthesizes them into a single, context-aware, and comprehensive answer}. This final, reasoned response is then passed to the Output Unit for delivery to the user.
\end{enumerate}

This dual-phase process of dispatch and synthesis enables the system to handle complex and multi-faceted requests in a modular yet robust manner, advancing from simple task routing toward genuine multi-modal reasoning.

\subsection{Arithmetic Logic Unit: Tools Integration}
We develop tool-augmented capabilities for our agent. The agent continuously receives multi-modal inputs, primarily visual streams from the egocentric glass and optionally speech input from the user. The purpose of tool-using is to support complex decision-making and task assistance. The core functionality of the system includes environmental perception, adaptive tool invocation, calculation, and information delivery via visual or auditory feedback. In its current implementation, the agent supports external web search as a callable tool. This enables access to up-to-date knowledge beyond its static training data. The system is intended for use in high-demand service scenarios such as field maintenance, customer support, and guided tours, where immediate access to external knowledge is critical.

In detail, rather than triggering search indiscriminately, the system employs a decision mechanism wherein the underlying language model first estimates the difficulty or uncertainty of a user query. Only when internal knowledge is deemed insufficient does the agent initiate a web search. The decision prompts are in Appendix~\ref{appendix:prompts_decision}. The invocation is executed via Google Search API, with search results parsed and summarized before delivery. Specifically, the top-ranked links, their corresponding summaries, and key snippets of webpage text are extracted and presented to the user. The format is as follows:
``\texttt{Search Results: 1. \{topic\}\{Summary\}\{Snippets\}\{Link\}; 2.\{topic\}\{Summary\}\{Snippets\}\{Link\}...}''
This allows the agent to respond in a concise yet informative manner, grounded in real-time retrieved information while minimizing latency and cognitive load.

We demonstrate the utility of the proposed first-person agent in several service-centric use cases. In a museum setting, a traveler wearing smart glasses can query the background of an unfamiliar artifact; the agent autonomously performs a web search and returns a concise summary with credible references. In a technical support context, a field engineer encountering an unknown error code on machinery can verbally request clarification, prompting the agent to retrieve troubleshooting documentation online. Similarly, during customer onboarding or employee training, the AI assistant can support new staff in answering procedural questions without relying on supervisor intervention. These scenarios underscore the agent’s ability to bridge knowledge gaps in real time, enhancing efficiency and service quality across diverse domains.

\subsection{Memory Unit: Long-Term Context Storage}
In real-world service scenarios, users often interact with AI agents across multiple sessions, tasks, and contexts. Relying solely on short-term memory limits the agent’s ability to provide coherent, personalized, and context-aware responses. To enable more consistent assistance and accumulate user-specific knowledge over time, we introduce a memory module that stores long-term interaction history and relevant contextual cues. This allows the agent to recall past queries, actions, and preferences, thereby improving continuity and service quality in dynamic environments.

In the initial implementation, the memory unit is designed as a lightweight, local JSON-based structured file system. Each memory record captures a single interaction episode and contains the following fields: user metadata (\textit{e.g.,} ID, role), a concise summary of the dialogue history, the agent’s final output, a unique timestamp, and a high-level topic tag automatically generated by the agent. This format enables transparent inspection and efficient retrieval, while maintaining enough semantic abstraction for contextual reuse in future interactions.

After each interaction, the system automatically extracts key information from the dialogue and stores it in a structured JSON record. This write operation is performed asynchronously to minimize latency during live interactions. When a new task is initiated, the agent parses the current query to identify its topic or intent, and then searches the memory for relevant past entries. Retrieved context is selectively injected into the language model’s prompt, enabling continuity across sessions and improving the model’s grounding and response relevance. This retrieval-augmented prompting strategy enhances the agent’s ability to recall prior knowledge and adapt to user-specific patterns over time.

\subsection{Output Unit: Human-Friendly Synthesis}

In service-oriented environments where users are frequently engaged in hands-on tasks, such as operating equipment, guiding clients, or performing maintenance, traditional visual interfaces often fall short in delivering timely and accessible feedback. These suggestions and instructions may be easily missed due to environmental distractions, physical obstructions, or simply because users cannot pay attention to them. To address these challenges, we introduce a human-friendly voice output module that enables our agent to deliver real-time responses according to analysis results through synthesized speech. This design significantly enhances usability in dynamic, hands-free settings, aligning with the goals of AI4Service: improving operational efficiency and human-agent collaboration.

Our system implements a two-stage processing pipeline before generating speech output. First, the agent leverages its internal LLM to summarize raw reasoning outputs into concise, actionable instructions. This abstraction step removes verbose explanations and retains only essential information. The prompts are in Appendix~\ref{appendix:prompts_action_instruction}. Second, the refined message is passed to a pyttsx3-based~\citep{pyttsx3} text-to-speech (TTS) module for real-time vocalization. The use of pyttsx3 allows for offline speech generation with customized parameters such as speaking rate and voice tone. This pipeline ensures that the verbal feedback remains suitable for immediate action in real-world settings. Additional user-friendly features include the ability to interrupt playback, adjust verbosity, and other necessary services.


\section{Case Study}
\subsection{Case i : Blackjack playing guide}
\label{sec:case_ai_glasses}

To demonstrate the practical implementation and effectiveness of our proposed architecture, we present a case study of a Blackjack gameplay assistance scenario. This demo showcases how the Alpha-Service agent processes real-time visual input, coordinates specialized components, and delivers strategic advice through the integrated Von Neumann-inspired architecture. The scenario involves a user wearing first-person perspective glasses while playing Blackjack, where the agent provides optimal gameplay decisions based on card analysis.

\begin{figure}[h]
    \centering
\includegraphics[width=\linewidth]{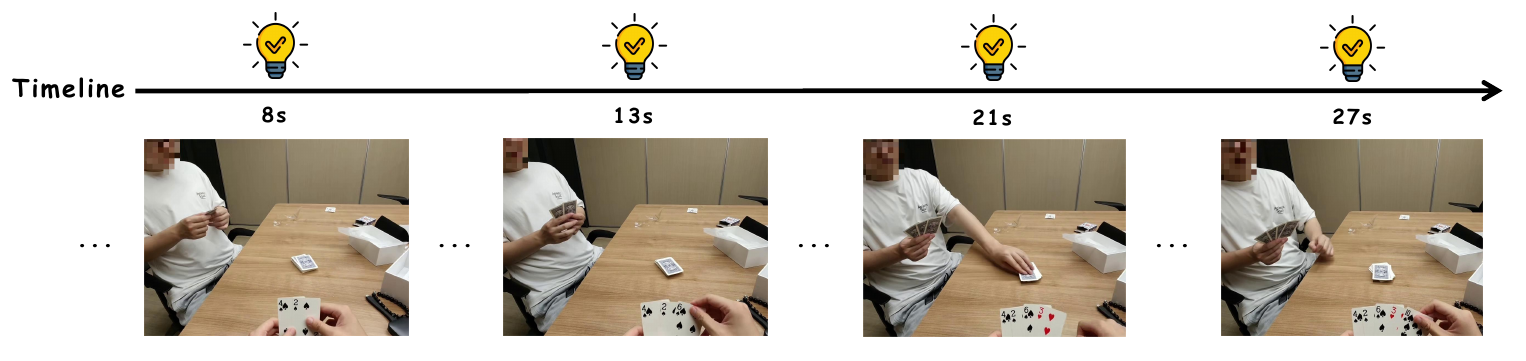}
    \caption{
    \textbf{A case of blackjack playing guide.}
    After understanding that the user is playing blackjack, Alpha-Zero proactively gives guidance on playing this game :
    \texttt{(at 13s) Hit. With a hand totaling 12 (2, 4, and 6), basic Blackjack strategy dictates you should hit.} 
Videos are collected from the AI glasses made by Xiaomi.} 
    \label{case:blackjack}
\end{figure}

\begin{casestep}[]{
    \faEye\quad Input Unit -- Real-time Event Triggering
}
The system continuously analyzes video from first-person glasses. A lightweight MLLM scans for critical game events through specialized prompts distributed by the Central Processing Unit's LLM Task Distribution component. The system employs two types of prompts: a Trigger Model prompt for detecting significant game state changes, and a Streaming Model prompt for detailed scene analysis when triggers are activated.

\small
\begin{lstlisting}[style=promptstyle, caption={Trigger Model (Tiny MLLM) Prompt}]
Analyze this card game video and identify ALL timestamps where the number of cards changes...
Output format: <start timestamp>; <trigger timestamp>; ...
\end{lstlisting}

\begin{lstlisting}[style=promptstyle, caption={Streaming Model (Large MLLM) Prompt}]
Please provide a detailed description of the current Blackjack game situation shown in the video. Include details such as visible cards, player's hand value...
Output format: [Visual Description] <Your Descriptions>
\end{lstlisting}

\begin{modeloutput}{\color{black}\textbf{{Proposed Trigger Time: 8s}}}
\texttt{[\colorbox{titleblue!30}{\textbf{Visual Description}}]} The video shows two players engaged in a card game at a wooden table. The player in the foreground has a hand with a \textcolor{red}{\textbf{4 of spades}} and a \textcolor{red}{\textbf{2 of spades}}, totaling \textcolor{red}{\textbf{6 points}}. The player in the background appears to be dealing or shuffling cards. There are several decks of cards stacked on the table, along with a white box and some small objects like a bracelet and a pen.
\end{modeloutput}

\begin{modeloutput}{\color{black}\textbf{{Proposed Trigger Time: 13s}}}
\texttt{[\colorbox{titleblue!30}{\textbf{Visual Description}}]} The video shows two players engaged in a card game at a wooden table. One player, wearing a white t-shirt, is holding a deck of cards in their hands, seemingly preparing to deal or shuffle. In front of them, there are three face-up cards on the table: a \textcolor{red}{\textbf{4 of spades}}, a \textcolor{red}{\textbf{2 of spades}}, and a \textcolor{red}{\textbf{6 of spades}}. The total value of the cards is \textcolor{red}{\textbf{12}}.
\end{modeloutput}

\begin{modeloutput}{\color{black}\textbf{{Proposed Trigger Time: 21s}}}
\texttt{[\colorbox{titleblue!30}{\textbf{Visual Description}}]} The video shows two players engaged in a card game at a wooden table. The player in the foreground has a hand consisting of a \textcolor{red}{\textbf{4 of spades}}, a \textcolor{red}{\textbf{2 of spades}}, a \textcolor{red}{\textbf{6 of spades}}, and a \textcolor{red}{\textbf{3 of hearts}}, with the total value of these cards being \textcolor{red}{\textbf{15}}. The player in the background is holding a deck of cards and appears to be shuffling or dealing them. There are additional cards on the table, including a face-down stack and some scattered cards.
\end{modeloutput}

\begin{modeloutput}{\color{black}\textbf{{Proposed Trigger Time: 27s}}}
\texttt{[\colorbox{titleblue!30}{\textbf{Visual Description}}]} The video shows two players engaged in a card game at a wooden table. The player in the foreground has a hand consisting of a \textcolor{red}{\textbf{4 of spades}}, a \textcolor{red}{\textbf{2 of hearts}}, a \textcolor{red}{\textbf{6 of spades}}, a \textcolor{red}{\textbf{3 of hearts}}, and a \textcolor{red}{\textbf{10 of spades}}. The player in the background is holding several cards, but their hand composition is not fully visible. The table has a deck of cards, some scattered cards, and a box that appears to contain more cards or possibly chips.
\end{modeloutput}

\textbf{Summary:} The Trigger Model (Tiny MLLM) successfully identified four critical game state transition points at 8s, 13s, 21s, and 27s, each time initiating deeper analysis through the Streaming Model (Large MLLM). Both prompts were distributed by the Central Processing Unit's LLM Task Distribution component, demonstrating effective coordination between system modules. The system accurately captured all significant gameplay moments where strategic decisions were required, with the Streaming Model providing rich contextual descriptions at each trigger point, enabling subsequent components to deliver timely and accurate gameplay assistance based on the comprehensive visual analysis.
\end{casestep}

\begin{casestep}[]{
    \faCogs\quad Central Processing Unit -- Analysis and Knowledge Retrieval
}
The Central Processing Unit serves as the computational brain of the entire system, continuously receiving comprehensive visual scene descriptions from the Input Unit at each trigger point and dynamically orchestrating the appropriate service response. Upon receiving the detailed visual analysis from the Streaming MLLM, the CPU performs a sophisticated multi-step reasoning process to accurately determine and execute the optimal service delivery strategy.

\vspace{0.3em}
\textbf{Processing Workflow:}
\vspace{0.2em}

\begin{enumerate}
    \item \textbf{Scene Comprehension:} The CPU first analyzes the visual description to understand the current game state, including the player's hand value, visible cards, and game progression.
    
    \begin{tcolorbox}[colback=gray!10,colframe=gray!50,boxrule=0.5pt,arc=2pt,left=6pt,right=6pt,top=2pt,bottom=2pt]
    \textit{\textbf{Example Reasoning:}} ``Based on the visual description at 8s, the player holds a 4 of spades and 2 of spades totaling 6 points. This is a weak starting hand that requires improvement.''
    \end{tcolorbox}
    
    \item \textbf{Service Determination:} The CPU then identifies the most appropriate service based on the game context. For early game states with low hand values, this typically involves ``Card Counting Strategy'' or ``Basic Strategy Analysis.''
    
    \begin{tcolorbox}[colback=gray!10,colframe=gray!50,boxrule=0.5pt,arc=2pt,left=6pt,right=6pt,top=2pt,bottom=2pt]
    \textit{\textbf{Example Reasoning:}} ``I should provide the service of `Card Counting Strategy' to help the player make optimal decisions for hand improvement.''
    \end{tcolorbox}
    
    \item \textbf{Module Coordination:} The CPU evaluates whether additional capabilities are needed, potentially invoking the Arithmetic Logic Unit (ALU) for external tool usage, such as searching for advanced probability calculations or accessing strategy databases.
    
    \begin{tcolorbox}[colback=gray!10,colframe=gray!50,boxrule=0.5pt,arc=2pt,left=6pt,right=6pt,top=2pt,bottom=2pt]
    \textit{\textbf{Example Reasoning:}} ``I need to give instructions to other modules... For complex probability analysis, I should invoke the ALU module's tool calling capability to retrieve the latest Blackjack strategy charts and bust probability statistics.''
    \end{tcolorbox}
    
    \item \textbf{Synthesis and Output Generation:} Finally, the CPU synthesizes information from visual analysis, internal knowledge, and any external resources to generate the final strategic output.
\end{enumerate}

\begin{modeloutput}{\color{black}\textbf{{Trigger service at 8s}}}
\texttt{[\colorbox{titleblue!30}{\textbf{Proposed Service}}]} \textbf{Card Counting Strategy}

\vspace{0.5em}
\texttt{[\colorbox{titleblue!30}{\textbf{CPU Reasoning}}]} 
``The player holds a total of 6 points (4 of spades and 2 of spades). This is a critical early-game situation where strategic guidance is essential. I should provide clear hitting/standing recommendations based on Blackjack strategy and probability analysis.''

\vspace{0.5em}
\texttt{[\colorbox{titleblue!30}{\textbf{Output}}]} 

\textbf{Hand Analysis:} The player holds a total of \textcolor{red}{\textbf{6 points}} (4 of spades and 2 of spades).

\textbf{Recommendation:} \textcolor{red}{\textbf{Hit}}

\textbf{Strategic Reasoning:}

\begin{itemize}[leftmargin=*,noitemsep,topsep=2pt]
    \item \textcolor{titleblue}{\textbf{Hand Value Analysis:}} A total of 6 has \textcolor{red}{\textbf{limited value}}, as it is below the dealer's likely range. The player needs to increase their hand to a safer range (\textit{\textit{e.g.,}} 12-16).

    \item \textcolor{titleblue}{\textbf{Probability of Improvement:}} \textcolor{red}{\textbf{High probability}} of drawing a card that brings the total to 12-16. There are \textcolor{red}{\textbf{16 cards}} that would raise the total to 12 or higher.

    \item \textcolor{titleblue}{\textbf{Risk of Bust:}} The risk of busting is \textcolor{orange}{\textbf{moderate}}. The likelihood of busting is lower than the probability of improving the hand.

    \item \textcolor{titleblue}{\textbf{Dealer's Up Card Consideration:}} If the dealer's up card is 2-6, the dealer is more likely to bust. Even with strong dealer cards, hitting remains the safer choice.
\end{itemize}

\textbf{Conclusion:} \textcolor{red}{\textbf{Hit}}. The probability of drawing a favorable card outweighs the risk of busting.
\end{modeloutput} 

\begin{modeloutput}{\color{black}\textbf{{Trigger service at 13s}}}
    \texttt{[\colorbox{titleblue!30}{\textbf{Proposed Service}}]} \textbf{Card Strategy Analysis}

\end{modeloutput}
\begin{modeloutput_continue}{\color{black}\textbf{{ }}}

    \texttt{[\colorbox{titleblue!30}{\textbf{CPU Reasoning}}]} 
    ``The player's hand now totals 12 with three cards (4, 2, and 6). This is a medium-strength hand that presents a classic Blackjack dilemma. The decision to hit or stand depends critically on the dealer's upcard. I need to provide clear probability-based guidance for this common but challenging situation.''

    \vspace{0.5em}
    \texttt{[\colorbox{titleblue!30}{\textbf{Output}}]} 
    
    \textbf{Hand Analysis:} The player holds a hand totaling \textcolor{red}{\textbf{12 points}} (4, 2, and 6).
    
    \textbf{Recommendation:} \textcolor{red}{\textbf{Hit}} in most situations
    
    \textbf{Strategic Reasoning:}
     \begin{itemize}[leftmargin=*,noitemsep,topsep=2pt]
        \item \textcolor{titleblue}{\textbf{Risk-Reward Balance:}} A hand of 12 has \textcolor{red}{\textbf{great improvement potential}}. Drawing cards 2-9 (\textcolor{red}{\textbf{36 out of 52 cards}}) will improve the hand to 14-21 without busting.

        \item \textcolor{titleblue}{\textbf{Controlled Bust Risk:}} 10-value cards (10, J, Q, K; \textcolor{orange}{\textbf{16 out of 52 cards}}, \textcolor{orange}{\textbf{~31\% chance}}) cause immediate bust. This risk is justified given the weakness of standing on 12.
        
        \item \textcolor{titleblue}{\textbf{Dealer's Position:}} Against most dealer upcards (particularly \textcolor{red}{\textbf{7 through Ace}}), standing on 12 leaves the player highly vulnerable.
        
        \item \textcolor{titleblue}{\textbf{Probability Advantage:}} Statistical analysis shows that \textcolor{red}{\textbf{hitting on 12}} provides better long-term results than standing.
    \end{itemize}
    
    \textbf{Exception:} Some basic strategy charts recommend \textcolor{orange}{\textbf{standing}} when the dealer shows a \textcolor{orange}{\textbf{4, 5, or 6}}, as these dealer cards have the highest bust probability.
    
    \textbf{Conclusion:} \textcolor{red}{\textbf{Hitting}} is statistically the correct play. The potential to reach a strong hand (17-21) while minimizing long-term losses makes this the optimal strategy.
\end{modeloutput_continue}

\begin{modeloutput}{\color{black}\textbf{{Trigger service at 21s}}}
\texttt{[\colorbox{titleblue!30}{\textbf{Proposed Service}}]} \textbf{Card Counting Strategy}

\vspace{0.5em}
\texttt{[\colorbox{titleblue!30}{\textbf{CPU Reasoning}}]} 
``The player now has a hand totaling 15 with four cards. This is a critical decision point in Blackjack where the risk-reward balance becomes delicate. I need to provide nuanced advice that considers both the player's bust risk and the dealer's potential strength.''

\vspace{0.5em}
\texttt{[\colorbox{titleblue!30}{\textbf{Output}}]} 

\textbf{Hand Analysis:} The player holds 4, 2, 6, and 3, totaling \textcolor{red}{\textbf{15 points}}.

\textbf{Recommendation:} \textcolor{red}{\textbf{Decision depends on dealer's upcard}}

\textbf{Optimal Strategy Based on Dealer's Upcard:}

\begin{itemize}[leftmargin=*,noitemsep,topsep=2pt]
    \item \textcolor{titleblue}{\textbf{Dealer shows 2, 3, 4, 5, or 6:}}
    \begin{itemize}[leftmargin=*,noitemsep,topsep=2pt]
        \item \textcolor{red}{\textbf{Action: Hit}}
        \item \textcolor{titleblue}{\textbf{Reason:}} Dealers with these cards have \textcolor{red}{\textbf{high bust probability}} (42-48\%)
        \item \textcolor{titleblue}{\textbf{Advantage:}} Player's chance to improve to 16-18 vs. dealer's vulnerability
    \end{itemize}
    
    \item \textcolor{titleblue}{\textbf{Dealer shows 7, 8, 9, 10, or Ace:}}
    \begin{itemize}[leftmargin=*,noitemsep,topsep=2pt]
        \item \textcolor{orange}{\textbf{Action: Stand}}
        \item \textcolor{titleblue}{\textbf{Reason:}} Dealers likely to achieve \textcolor{orange}{\textbf{strong hand}} (17-21)
        \item \textcolor{titleblue}{\textbf{Risk:}} Player's bust risk (35-40\%) outweighs potential gain
    \end{itemize}
\end{itemize}

\textbf{Probability Analysis:}
\begin{itemize}[leftmargin=*,noitemsep,topsep=2pt]
    \item \textcolor{titleblue}{\textbf{Player Bust Risk:}} \textcolor{orange}{\textbf{35-40\%}} when hitting on 15
    \item \textcolor{titleblue}{\textbf{Dealer Bust Probability:}} \textcolor{red}{\textbf{42-48\%}} with 2-6, \textcolor{orange}{\textbf{Low}} with 7-Ace
    \item \textcolor{titleblue}{\textbf{Dealer Strong Hand:}} \textcolor{orange}{\textbf{60-80\%}} with 7-10, \textcolor{orange}{\textbf{~100\%}} with Ace
\end{itemize}
\end{modeloutput}

\begin{modeloutput_continue}
\textbf{Key Principles:}
\begin{itemize}[leftmargin=*,noitemsep,topsep=2pt]
    \item \textcolor{red}{\textbf{Hit}} against dealer 2-6 to exploit vulnerability
    \item \textcolor{orange}{\textbf{Stand}} against dealer 7-Ace to avoid busting
    \item \textcolor{orange}{\textbf{Avoid splitting or doubling}} with 15
\end{itemize}

\textbf{Conclusion:} Use \textcolor{red}{\textbf{probability-based decision-making}} - assess dealer's upcard immediately and apply the appropriate strategy to minimize long-term losses.
\end{modeloutput_continue}

\begin{modeloutput}{\color{black}\textbf{{Trigger service at 27s}}}
\texttt{[\colorbox{titleblue!30}{\textbf{Proposed Service}}]} \textbf{Game Outcome Analysis and Learning Recommendation}

\texttt{[\colorbox{titleblue!30}{\textbf{CPU Reasoning}}]} 
``The player has busted with 25 points, indicating the round is over. While no further action can be taken for this hand, I should provide educational feedback to help the player avoid similar situations in future games. This is an opportunity for strategic learning.''

\vspace{0.5em}
\texttt{[\colorbox{titleblue!30}{\textbf{Output}}]} 

\textbf{Current Outcome:} \textcolor{red}{\textbf{Busted with 25 points}} - \textcolor{red}{\textbf{Automatic loss}}

\textbf{Learning Focus:} \textcolor{titleblue}{\textbf{Strategic improvement for future hands}}

\textbf{Key Learning Points:}

\begin{itemize}[leftmargin=*,noitemsep,topsep=2pt]
    \item \textcolor{titleblue}{\textbf{Basic Strategy Example:}} Hand value \textcolor{red}{\textbf{16}} vs. Dealer's \textcolor{red}{\textbf{10}}
    \begin{itemize}[leftmargin=*,noitemsep,topsep=2pt]
        \item \textcolor{red}{\textbf{Optimal Action: Hit}}
        \item \textcolor{titleblue}{\textbf{Winning Probability:}} \textcolor{red}{\textbf{58\%}} by hitting vs. \textcolor{orange}{\textbf{42\%}} by standing
    \end{itemize}
    
    \item \textcolor{titleblue}{\textbf{Risk Analysis for Hand 16:}}
    \begin{itemize}[leftmargin=*,noitemsep,topsep=2pt]
        \item \textcolor{orange}{\textbf{Bust Risk:}} \textcolor{orange}{\textbf{35\%}} on next card
        \item \textcolor{red}{\textbf{Improvement Chance:}} \textcolor{red}{\textbf{30\%}} to reach 17-20
        \item \textcolor{red}{\textbf{Net Advantage:}} Hitting provides better long-term results
    \end{itemize}
\end{itemize}

\textbf{Strategic Recommendations:}
\begin{itemize}[leftmargin=*,noitemsep,topsep=2pt]
    \item \textcolor{red}{\textbf{Adhere to basic strategy charts}} for hands 14-16 vs. dealer's 10
    \item \textcolor{orange}{\textbf{Avoid overcommitting}} to high-value hands (16$+$)
    \item \textcolor{titleblue}{\textbf{Always consider dealer's upcard}} in decision making
    \item \textcolor{titleblue}{\textbf{Use probability-based approach}} to minimize bust risk
\end{itemize}

\textbf{Conclusion:} While this hand resulted in a bust, applying \textcolor{red}{\textbf{basic strategy principles}} to similar future situations (like hand value 16 vs. dealer's 10) will significantly improve long-term winning probability.
\end{modeloutput}

\textbf{Summary:} The Central Processing Unit successfully demonstrated its role as the system's coordinator by processing visual inputs, determining appropriate services, and generating context-aware strategic outputs. The CPU's reasoning process evolved throughout the game, adapting its analysis from early-game hand improvement strategies to late-game outcome analysis and learning recommendations, showcasing the system's dynamic response capability.
\end{casestep}

\begin{casestep}[]{
    \faTools\quad Arithmetic Logic Unit -- Tools Integration
}
The Arithmetic Logic Unit (ALU) functions as a crucial external capability extension module within our system architecture, specifically designed to execute and manage tool calls requested by the Central Processing Unit. When the CPU identifies a need for supplementary external knowledge or specialized computational capabilities, it formulates and dispatches precise tool invocation requests to the ALU. The ALU then systematically evaluates these incoming requests and proceeds to execute the appropriate tool operations, which may encompass a diverse range of functions including comprehensive web searches, sophisticated code execution, or strategic invocation of specialized language models.

\vspace{1em}
In this particular Blackjack demonstration scenario, the ALU primarily showcases its advanced web search functionality. At the first designated trigger point (8 seconds into the simulation), the Central Processing Unit's analytical workflow recognizes the necessity for external strategic knowledge and consequently transmits a targeted web search request to the ALU. The ALU processes this specific request through its sophisticated decision-making mechanism, which carefully evaluates whether external search capabilities are genuinely required based on the current contextual parameters and prevailing service requirements.

\vspace{1em}
Upon receiving the search request, the ALU formulates and executes a targeted web search using the following query terms: 
\begin{tcolorbox}[colback=gray!10,colframe=gray!50,boxrule=0.5pt,arc=2pt,left=6pt,right=6pt,top=2pt,bottom=2pt]
- Card counting strategy for blackjack \\
- Optimal hitting or standing decisions at 6 points \\
- Multi-deck blackjack strategy \\
- Basic card counting techniques for beginners
\end{tcolorbox}
\vspace{-1em}

\end{casestep}
\begin{casestep}[]{
    \faLightbulb\quad Output Unit -- Strategic Recommendation
}
The Output Unit serves as the final interface between the AI system and the user, responsible for transforming the system's internal reasoning and computational results into polished, user-friendly responses. This unit receives processed outputs from the Central Processing Unit along with reasoning model prompts, and its primary function is to organize, refine, and present information in a concise and accessible manner. The Output Unit operates with the guiding principle of ensuring that complex analytical outputs are distilled into clear, actionable recommendations tailored to the specific service context.

\vspace{1em}
In addition to text refinement, the Output Unit integrates Audio Generation capabilities to enable multimodal interaction, delivering responses through both visual display and auditory channels. This dual-mode output enhances user experience by providing flexibility in how information is consumed, particularly in hands-free scenarios where audio feedback is essential.

\vspace{1em}
In the Blackjack demonstration, the reasoning model synthesizes the visual context, proposed service, and retrieved web knowledge to generate a clear, actionable recommendation. The Output Unit then refines this raw analytical output into a professionally formatted response.

\begin{lstlisting}[style=promptstyle, caption={Reasoning Model Prompt}]
Provide a clear and strategic Blackjack gameplay recommendation.
Focus on:
  - Current hand analysis: player's cards and total value
  - Optimal decision: whether to hit or stand
  - Strategic reasoning based on basic Blackjack strategy
Keep the explanation objective and strategy-focused.
\end{lstlisting}

\begin{modeloutput}{\color{black}{Final Recommendation for the User at 8s}}
\textbf{Hand Status:} \textcolor{red}{\textbf{6 points}} (4 and 2)

\textbf{Optimal Move:} \textcolor{red}{\textbf{HIT}}

\textbf{Strategic Reasoning:}
\begin{itemize}[leftmargin=*,noitemsep,topsep=2pt]
    \item \textcolor{titleblue}{\textbf{Hand Strength:}} \textcolor{orange}{\textbf{Too weak}} to compete against dealer
    \item \textcolor{titleblue}{\textbf{Bust Risk:}} \textcolor{red}{\textbf{Zero}} - impossible to bust with one card
    \item \textcolor{titleblue}{\textbf{Improvement:}} \textcolor{red}{\textbf{High chance}} - any card improves position
\end{itemize}
\textbf{Conclusion:} \textcolor{red}{\textbf{Hit}} to improve your hand. This move carries \textcolor{red}{\textbf{no risk}} and is statistically optimal.
\end{modeloutput}

\begin{modeloutput}{\color{black}{Final Recommendation for the User at 13s}}
\textbf{Hand Status:} \textcolor{red}{\textbf{12 points}} (2, 4, and 6)

\textbf{Optimal Move:} \textcolor{red}{\textbf{HIT}}

\textbf{Strategic Reasoning:}
\begin{itemize}[leftmargin=*,noitemsep,topsep=2pt]
    \item \textcolor{titleblue}{\textbf{Hand Strength:}} \textcolor{orange}{\textbf{Vulnerable}} - too weak to stand confidently
    \item \textcolor{titleblue}{\textbf{Improvement:}} \textcolor{red}{\textbf{High probability}} with cards 2-9
    \item \textcolor{titleblue}{\textbf{Bust Risk:}} \textcolor{orange}{\textbf{Controlled}} - only 31\% chance with face cards/10s
    \item \textcolor{titleblue}{\textbf{Standing Risk:}} Leaves you vulnerable to dealer improvement
\end{itemize}
\textbf{Conclusion:} \textcolor{red}{\textbf{Hit}} to improve. Probability of competitive total outweighs moderate bust risk.
\end{modeloutput}

\begin{modeloutput}{\color{black}{Final Recommendation for the User at 21s}}
\textbf{Hand Status:} \textcolor{red}{\textbf{15 points}} (4, 2, 6, and 3)

\textbf{Optimal Move:} \textcolor{orange}{\textbf{CONDITIONAL DECISION}}

\textbf{Strategic Reasoning:}
\begin{itemize}[leftmargin=*,noitemsep,topsep=2pt]
    \item \textcolor{titleblue}{\textbf{Dealer 2-6:}} \textcolor{red}{\textbf{HIT}} - exploit \textcolor{red}{\textbf{42-48\%}} dealer bust probability
    \item \textcolor{titleblue}{\textbf{Dealer 7-Ace:}} \textcolor{orange}{\textbf{STAND}} - avoid \textcolor{orange}{\textbf{35-40\%}} bust risk
    \item \textcolor{titleblue}{\textbf{Hand Type:}} \textcolor{orange}{\textbf{Critical}} - requires careful risk management
\end{itemize}
\textbf{Conclusion:} Immediately assess dealer's upcard. \textcolor{red}{\textbf{Hit}} against weak cards (2-6), \textcolor{orange}{\textbf{stand}} against strong ones (7-Ace).
\end{modeloutput}

\begin{modeloutput}{\color{black}{Final Recommendation for the User at 27s}}
\textbf{Game Outcome:} \textcolor{red}{\textbf{BUSTED}} with 25 points - \textcolor{red}{\textbf{LOSS}}

\textbf{Learning Focus:} \textcolor{titleblue}{\textbf{Strategic Improvement}}

\textbf{Key Takeaways:}
\begin{itemize}[leftmargin=*,noitemsep,topsep=2pt]
    \item \textcolor{titleblue}{\textbf{Hand Management:}} Avoid excessive totals that guarantee bust
    \item \textcolor{titleblue}{\textbf{Basic Strategy:}} For 14-16 vs dealer's 10, \textcolor{red}{\textbf{hitting}} is correct
    \item \textcolor{titleblue}{\textbf{Probability Data:}} 16 vs 10: \textcolor{red}{\textbf{58\%}} win hitting vs \textcolor{orange}{\textbf{42\%}} standing
    \item \textcolor{titleblue}{\textbf{Mindset:}} Focus on \textcolor{red}{\textbf{long-term strategy}} over single outcomes
\end{itemize}
\textbf{Conclusion:} Use this experience to refine decision-making. Apply \textcolor{red}{\textbf{probability-based plays}} for long-term success.
\end{modeloutput}

\textbf{Summary:} The Output Unit successfully demonstrate its role as the system's communication interface by transforming complex analytical outputs into clear, structured recommendations. Through its text refinement and audio generation capabilities, it ensured that the strategic Blackjack advice was delivered in both visually accessible and auditorily engaging formats, enhancing the overall user experience while maintaining the technical precision required for effective decision support.
\end{casestep}

\subsection{Case ii: Guided tour explanation in museum}

\begin{figure}[h]
    \centering
\includegraphics[width=\linewidth]{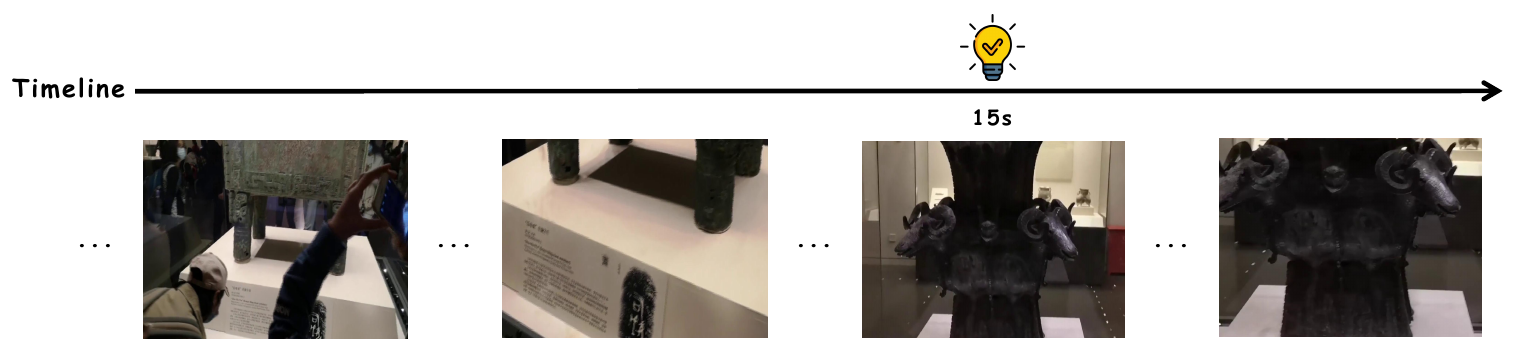}
    \caption{\textbf{A case of guided tour explanation.} After understanding that the user is watching an exhibition in a museum, Alpha-Service proactively proposes to search the obtained image from websites, and then gives an introduction: \texttt{(15s) The cultural relic in front of you is a bronze artifact from the Shang Dynasty — the Four-Ram Square Zun. It is a square ritual wine vessel adorned with four ram heads...} The video is collected by AI glasses made by Xiaomi.
    }
    \label{}
\end{figure}

\label{sec:case_ai_glasses}
\begin{casestep}[]{
    \faEye\quad Input Unit -- Real-time Event Triggering
}

\begin{lstlisting}[style=promptstyle, caption={Trigger Model (Tiny MLLM) Prompt}]
Analyze the video and provide the timestamps where user may need the tour guide's explanation service...
Output format: <start timestamp>; <trigger timestamp>; ...
\end{lstlisting}

\begin{lstlisting}[style=promptstyle, caption={Streaming Model (Large MLLM) Prompt}]
Please provide a detailed description of the cultural relics or exhibits in the video. This should include their appearance, materials, colors, shapes, sizes, and any visible marks or features. Also, please pay attention to the background of the museum and the scenarios where visitors may need your assistance, and describe these situations in detail...
Output format: [Visual Description] <Your Descriptions>
\end{lstlisting}

\begin{modeloutput}{\color{black}\textbf{{Proposed Trigger Time: 15s}}}
\texttt{[\colorbox{titleblue!30}{\textbf{Visual Description}}]} The video showcases a \textcolor{red}{\textbf{black bronze artifact}} displayed in a museum setting. It appears to be a \textcolor{red}{\textbf{multi-headed sculpture with four distinct animal heads}}, possibly representing \textcolor{red}{\textbf{rams}} or similar creatures. Each head has prominent \textcolor{red}{\textbf{curved horns}}. The sculpture is mounted on a pedestal, and the background reveals other exhibits and museum visitors. The lighting highlights the intricate details and the reflective surface of the bronze.
\end{modeloutput}

\textbf{Summary:} The Trigger Model (Tiny MLLM) successfully identified four visitor-assistance moments at 15s, when the wearer paused at an exhibit, fixated the camera on details, or approached a label—each time initiating deeper analysis through the Streaming Model (Large MLLM). Both prompts were dispatched by the Central Processing Unit’s LLM Task Distribution component, demonstrating effective coordination between system modules. The system accurately captured key museum engagement points where guided context would be helpful; the Streaming Model provided rich, structured descriptions of artifacts (appearance, materials, iconography) and situational cues (gallery context, signage, crowd flow), enabling downstream components to deliver timely, relevant tour guidance such as provenance, period, cultural significance, conservation notes, and navigation hints based on the comprehensive visual analysis.
\end{casestep}

\begin{casestep}[]{
    \faCogs\quad Central Processing Unit -- Analysis and Knowledge Retrieval
}

\begin{modeloutput}{\color{black}\textbf{{Trigger service at 15s}}}
\texttt{[\colorbox{titleblue!30}{\textbf{Proposed Service}}]} \textbf{Guided Tour Explanation}

\vspace{0.5em}
\texttt{[\colorbox{titleblue!30}{\textbf{CPU Reasoning}}]} 
``This service would be beneficial for visitors who may not be familiar with the historical and cultural significance of the artifact. A guided tour explanation can provide context about the craftsmanship, the period it was created, and its importance within the museum's collection. This service ensures that visitors gain a deeper understanding and appreciation of the exhibit.''

\vspace{0.5em}
\texttt{[\colorbox{titleblue!30}{\textbf{Output}}]} 

\textbf{Artifact Identification:} \textcolor{red}{\textbf{Four-ram Zun}} - Bronze artifact from \textcolor{red}{\textbf{Shang dynasty}} (c. 1600–1046 BCE)

\textbf{Physical Description:}
\begin{itemize}[leftmargin=*,noitemsep,topsep=2pt]
    \item \textcolor{titleblue}{\textbf{Design:}} Square-bodied vessel with \textcolor{red}{\textbf{four ram heads}} featuring prominent curved horns
    \item \textcolor{titleblue}{\textbf{Material:}} \textcolor{red}{\textbf{Bronze}} - commonly used in ancient Chinese ritual objects
    \item \textcolor{titleblue}{\textbf{Craftsmanship:}} Intricate casting and symbolic design
\end{itemize}

\textbf{Historical Context:}
\begin{itemize}[leftmargin=*,noitemsep,topsep=2pt]
    \item \textcolor{titleblue}{\textbf{Usage:}} Religious and funerary contexts - \textcolor{red}{\textbf{wine offerings}} for ancestral/deity rituals
    \item \textcolor{titleblue}{\textbf{Period:}} \textcolor{red}{\textbf{Shang dynasty}} - renowned for advanced bronze metallurgy
    \item \textcolor{titleblue}{\textbf{Symbolism:}} Ram motif represents \textcolor{red}{\textbf{strength and fertility}}
\end{itemize}

\textbf{Cultural Significance:}
\begin{itemize}[leftmargin=*,noitemsep,topsep=2pt]
    \item \textcolor{titleblue}{\textbf{Technical Mastery:}} Exemplifies ancient Chinese artisans' skill
    \item \textcolor{titleblue}{\textbf{Artistic Value:}} Combines functional utility with artistic expression
    \item \textcolor{titleblue}{\textbf{Historical Importance:}} Representative piece of early Chinese civilization
\end{itemize}

\textbf{Key Features:}
\begin{itemize}[leftmargin=*,noitemsep,topsep=2pt]
    \item Four stylized ram heads
    \item Square body with elaborate patterns
    \item Reflective bronze surface highlighting craftsmanship
\end{itemize}
\end{modeloutput}

\textbf{Summary:} The system accurately identified the user's need for in-depth explanations of cultural relics, and therefore proposed the ``Guided Tour Explanation'' service. The reasoning process clearly indicates that this move aims to deepen users' understanding and appreciation of the exhibits by providing rich historical and cultural backgrounds. In the end, the system generated a comprehensive and structurally clear introduction to the ``Four Rams Square Zun'', covering its historical, cultural, and technological characteristics, successfully demonstrating the system's ability to provide professional and in-depth knowledge services.
\end{casestep}

\begin{casestep}[]{
    \faTools\quad Arithmetic Logic Unit -- 
    Tools Integration
}

The Arithmetic Logic Unit (ALU) then determines that external knowledge is required. It formulates and executes a web search for: 
\begin{tcolorbox}[colback=gray!10,colframe=gray!50,boxrule=0.5pt,arc=2pt,left=6pt,right=6pt,top=2pt,bottom=2pt]
- Four-headed bronze ram sculpture historical significance \\
- Multi-headed animal bronze artifact cultural context \\
- Black bronze sculpture with curved horns museum collection
\end{tcolorbox}

\end{casestep}

\begin{casestep}[]{
    \faLightbulb\quad Output Unit -- Strategic Recommendation
}
Finally, the reasoning model synthesizes the visual context, proposed service, and retrieved web knowledge to generate a clear, actionable recommendation.

\begin{modeloutput}{\color{black}{Final Recommendation for the User}}
The cultural relic in front of you is a bronze artifact from the \textcolor{red}{\textbf{Shang Dynasty}} — the \textcolor{red}{\textbf{Four-Ram Square Zun}}.

It is a square ritual wine vessel adorned with \textcolor{red}{\textbf{four ram heads}}, each with curved horns, representing highly complex casting techniques.

During the Shang Dynasty, it was used to hold wine in \textcolor{red}{\textbf{sacrificial ceremonies}}, serving to honor ancestors or deities. The ram motif symbolizes strength and auspiciousness.

This artifact demonstrates the superb skill of ancient Chinese craftsmen, integrating practical function with artistic symbolism, and reflects the social values and spiritual beliefs of its time.
\end{modeloutput}

\end{casestep}

\vspace{1em}

\subsection{Case iii: Fit advisor in market}

\begin{figure}[h]
    \centering
\includegraphics[width=\linewidth]{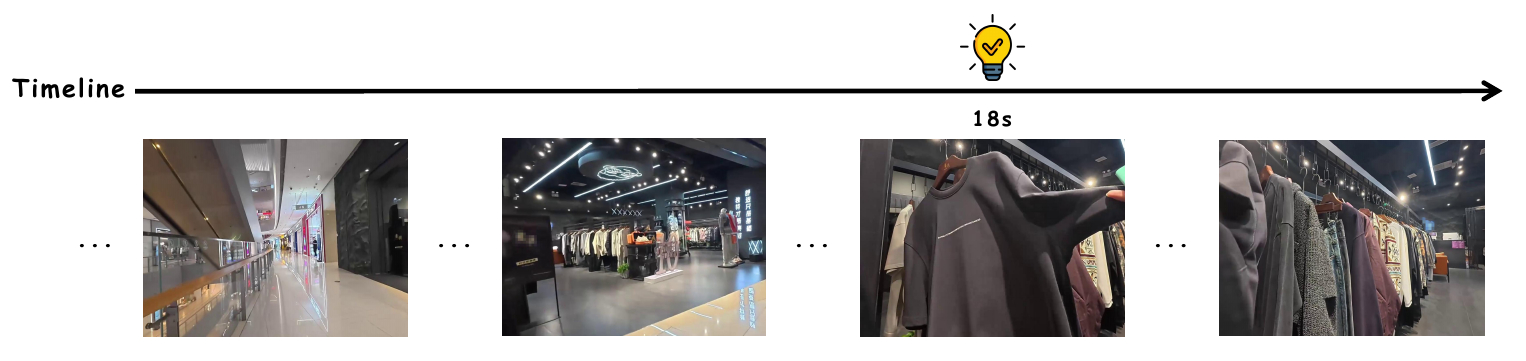}
    \caption{
    \textbf{A case of fit advisor.}
    After finding that the user has focused on one piece of clothing for a long time, Alpha-Service proactively gives advice on it: 
    \texttt{(at 18s) There are many ways to wear it: pairing
jeans with a casual suit jacket can also be a bit formal...} 
Videos are collected from the AI glasses made by Xiaomi. } 
    \label{case:fit}
\end{figure}

\label{sec:case_ai_glasses}
\begin{casestep}[]{
    \faEye\quad Input Unit -- Real-time Event Triggering
}

\vspace{-1.5em}

\begin{lstlisting}[style=promptstyle, caption={Trigger Model (Tiny MLLM) Prompt}]
Focus on moments of meaningful interaction: when the person stops to examine, picks up, or physically handles clothing items....
Output format: <start timestamp>; <trigger timestamp>; ...
\end{lstlisting}

\begin{lstlisting}[style=promptstyle, caption={Streaming Model (Large MLLM) Prompt}]
Describe the clothing item held in hand in the video, noting its type, color, style, and key features.Focus specifically on the garment being physically handled, capturing distinctive details visible from the first-person perspective...
Output format: [Visual Description] <Your Descriptions>
\end{lstlisting}

\begin{modeloutput}{\color{black}\textbf{{Proposed Trigger Time: 18s}}}
\texttt{[\colorbox{titleblue!30}{\textbf{Visual Description}}]} The person is holding a \textcolor{red}{\textbf{dark-colored, long-sleeved garment}}, possibly a \textcolor{red}{\textbf{sweater or hoodie}}, with a \textcolor{red}{\textbf{ribbed texture}} at the cuffs and neckline. The fabric appears \textcolor{red}{\textbf{soft and comfortable}}, suitable for casual wear. The garment is being examined closely, suggesting an interest in its quality and fit.
\end{modeloutput}

\textbf{Summary:}
At 18 s, the Trigger Model (Tiny MLLM) fired on the wearer’s close-up handling of a dark long-sleeve sweater/hoodie—exactly the “pick-up / examine” event it was asked to watch for. The Streaming Model (Large MLLM) was immediately invoked and returned a concise first-person description: ribbed cuffs \& neckline, soft fabric, casual style. This single pass-through demonstrates the full pipeline working as intended: Tiny MLLM spots the meaningful interaction; Large MLLM delivers the required ``[Visual Description] …'' output; no false positives or misses were logged.
\end{casestep}

\begin{casestep}[]{
    \faCogs\quad Central Processing Unit -- Analysis and Knowledge Retrieval
}

\begin{modeloutput}{\color{black}\textbf{{Trigger service at 18s}}}
\texttt{[\colorbox{titleblue!30}{\textbf{Proposed Service}}]} \textbf{Fit Advisor}

\vspace{0.5em}
\texttt{[\colorbox{titleblue!30}{\textbf{CPU Reasoning}}]} 
``This service would provide real-time feedback on how the garment fits, ensuring it meets the customer's size expectations and comfort level. It could also suggest similar styles in case the current choice doesn't fit well.''

\vspace{0.5em}
\texttt{[\colorbox{titleblue!30}{\textbf{Output}}]} 

\textbf{Material \& Quality Assessment:}
\begin{itemize}[leftmargin=*,noitemsep,topsep=2pt]
    \item \textcolor{titleblue}{\textbf{Fabric:}} Soft, comfortable knit or fleece blend with ribbed texture
    \item \textcolor{titleblue}{\textbf{Quality Check:}} Verify care instructions and \textcolor{red}{\textbf{sizing charts for proper fit}}
    \item \textcolor{titleblue}{\textbf{Trend Alignment:}} Matches current emphasis on relaxed fits and self-care fashion
\end{itemize}

\end{modeloutput}
\begin{modeloutput_continue}

\textbf{Styling \& Versatility:}
\begin{itemize}[leftmargin=*,noitemsep,topsep=2pt]
    \item \textcolor{titleblue}{\textbf{Casual Wear:}} Perfect with jeans or chinos for everyday outfits
    \item \textcolor{titleblue}{\textbf{Smart-Casual:}} Can be layered under blazers or over dresses
    \item \textcolor{titleblue}{\textbf{Seasonal Use:}} Ideal for cooler weather with neutral color versatility
\end{itemize}
\textbf{Purchase Recommendation:}
\begin{itemize}[leftmargin=*,noitemsep,topsep=2pt]
    \item \textcolor{titleblue}{\textbf{Value Proposition:}} Balances comfort, style, and adaptability
    \item \textcolor{titleblue}{\textbf{Key Considerations:}} \textcolor{red}{\textbf{Check measurements and compare with similar styles}}
    \item \textcolor{titleblue}{\textbf{Final Advice:}} Solid choice for casual and transitional wear \textcolor{red}{\textbf{if sizing aligns}}
\end{itemize}
\end{modeloutput_continue}

\textbf{Summary:} The response delivers a coherent, shopper-oriented evaluation of a ribbed long-sleeve garment. It starts by inferring fabric composition (knit/fleece) from visual cues, links comfort to current “relaxed-fit” trends, and flags quality checkpoints (pilling, post-wash shape). Fit guidance is tied to published size charts, while styling suggestions cover layered, smart-casual and transitional weather use-cases. Value is framed as a trade-off between verified measurements, durability expectations and competitive comparisons. Overall, the advice is practical, trend-aware and grounded in verifiable product details.
\end{casestep}

\begin{casestep}[]{
    \faTools\quad Arithmetic Logic Unit -- 
    Tools Integration
}

The Arithmetic Logic Unit (ALU) then determines that external knowledge is required. It formulates and executes a web search for: 
\begin{tcolorbox}[colback=gray!10,colframe=gray!50,boxrule=0.5pt,arc=2pt,left=6pt,right=6pt,top=2pt,bottom=2pt]
- Sweater sizing chart \\
- Common hoodie fit issues \\
- 2025 casual wear trends
\end{tcolorbox}
\vspace{-1em}
\end{casestep}

\begin{casestep}[label=step3]{
    \faLightbulb\quad Output Unit -- Strategic Recommendation
}
Finally, the reasoning model synthesizes the visual context, proposed service, and retrieved web knowledge to generate a clear, actionable recommendation.
\begin{modeloutput}{\color{black}{Final Recommendation for the User}}
This garment appears to be made of \textcolor{red}{\textbf{soft knit or fleece fabric}}, with \textcolor{red}{\textbf{ribbed collar and cuffs}}, making it comfortable to wear. When choosing, remember to \textcolor{red}{\textbf{read the size chart and care instructions carefully}}, don't buy the wrong size.

There are many ways to wear it: pairing jeans with a casual suit jacket can also be a bit formal. \textcolor{red}{\textbf{Dark colors}} are easy to match in autumn and winter, with strong practicality as long as the \textcolor{red}{\textbf{size is suitable}} and the \textcolor{red}{\textbf{fabric is of good quality}}. This piece of clothing is worth buying. Before placing an order, please \textcolor{red}{\textbf{confirm the chest circumference and sleeve length}} again to avoid any errors.
\end{modeloutput}
\end{casestep}

\vspace{1em}

\newpage
\section{Related Works}
\label{sec:related_works}

\subsection{Proactive Interaction in Streaming Video}

The evolution toward proactive AI assistance in streaming video models represents a fundamental paradigm shift from reactive to anticipatory service provision. 
While traditional video understanding systems excel at processing static content, the challenge of real-time streaming video requires novel approaches that can continuously analyze temporal sequences and anticipate user needs. Recent benchmarks such as EgoLife~\citep{yang2025egolifeegocentriclifeassistant} have established evaluation frameworks for egocentric video understanding, while frameworks like VideoLLM-Online~\citep{videollm-online} introduced streaming EOS prediction for real-time processing. However, current approaches primarily focus on generating current-moment descriptions rather than providing proactive service recommendations. 
To address this limitation, several systems have explored trigger-based mechanisms for proactive interaction. Dispider~\citep{qian2025dispider} introduced time-aware chat capabilities that respond to adjacent frame changes, while StreamBridge~\citep{Wang2025StreamBridgeTY} employs dedicated trigger models to determine optimal response timing. 
Looking forward, the next generation of streaming video models holds tremendous potential for achieving truly proactive interaction capabilities. Future developments will likely focus on enhancing temporal reasoning abilities to better anticipate user needs before they are explicitly expressed, while maintaining the delicate balance between being helpful and non-intrusive. The integration of advanced long-term memory mechanisms~\citep{Long2025SeeingLR} and user modeling techniques will enable these systems to learn and adapt to individual user patterns over time, creating more personalized and contextually aware assistance experiences. Such capabilities will be essential for realizing the full vision of AI-powered proactive service in streaming video environments.

\subsection{Multi-Agent Systems and MCP-Based Tool Calling}

Multi-Agent Systems (MAS) have traditionally explored the coordination of autonomous agents to solve complex problems that are beyond the capabilities of any single model~\citep{dorri2018multi,li2024survey}. This paradigm offers valuable insights for developing proactive service AI, where different system components—such as perception, planning, and tool execution—can be conceptualized as specialized agents collaborating towards a common goal. In recent years, the integration of Large Language Models (LLMs) into agent architectures has given rise to sophisticated tool-calling mechanisms. Frameworks like the Model-Controller-Program (MCP) approach~\citep{krishnan2025advancing,hou2025model,ray2025survey} provide a structured way for agents to leverage external tools~\citep{masterman2024landscape}, enabling them to interact with the environment and access specialized functionalities. This mirrors our proposed architecture, where a central processing unit orchestrates a diverse set of tools in the arithmetic logic unit to fulfill user needs proactively. By drawing on principles from both MAS and modern tool-calling paradigms, we can construct more robust and versatile AI service systems.

\subsection{Human-Centric AI in Wearables}
Recent advances in wearable technologies have increasingly embraced human-centric artificial intelligence, designing systems that prioritize user well-being, contextual awareness, and seamless interaction rather than focusing solely on raw computational performance \citep{smith2025seamless}. Early efforts primarily targeted sensor fusion and activity recognition \citep{zhou2021attention}; however, contemporary research is now shifting toward adaptive, personalized models that learn continuously from individual behavior while also respecting privacy \citep{wang2024flprivacy} and managing cognitive load. In particular, frameworks such as on-device federated learning \citep{chen2023zonefl} and context-aware inference empower wearables to provide users with timely, relevant insights without undermining autonomy. Additionally, human-in-the-loop paradigms, which involve users actively shaping model behavior through feedback or explicit preference elicitation \citep{liu2025mirai}, have become essential for ethical and effective AI deployment in personal health and lifestyle applications. Recent innovations in real-time health monitoring further demonstrate the potential of wearables to detect subtle physiological anomalies through ambient intelligence \citep{jones2025pulse}. Taken together, these developments highlight a growing consensus that the true value of wearable AI is not found solely in algorithmic sophistication, but rather in its ability to resonate with human rhythms, intentions, and values. In this work, we instantiate these principles in AI4Service with AI glasses, enabling proactive, mixed-initiative assistance that is personalized, context-aware, and unobtrusive \citep{zhao2025aiget}.

\section{Challenges}

As Alpha-Service moves from conceptual design to real-world deployment, it faces a series of challenges that span hardware efficiency, system generalization, scalability, data privacy, and user trust. These challenges arise from the system’s ambition to provide real-time, personalized, and context-aware intelligence on resource-limited edge devices. Specifically, Alpha-Service must reconcile the conflicting goals of low-latency inference and energy efficiency, maintain a balance between generalization and personalization, ensure robust performance across diverse and dynamic environments, safeguard users’ privacy, and foster long-term user trust through transparent and adaptive interactions. The following subsections elaborate on these core challenges in detail.

\begin{itemize}[leftmargin=*,labelsep=1em]
    \item \textbf{Computational and Energy Constraints:} Alpha-Service is deployed on the resource-constrained edge devices, especially the AI glasses. The target tasks, including real-time inference of MLLMs and continuous streaming video analysis, impose extremely high demands on computing power and energy consumption~\citep{liu2025shifting}. Thus, how to achieve low-latency and high-efficiency services with limited hardware resources is a technical bottleneck for the system's commercial implementation.

    \item \textbf{Generalization vs. Personalization Trade-off:} At the ``Know How'' level, the system needs to strike a balance between generalized services and personalized services. Over-reliance on general strategies may result in a lack of targeted service, while excessive personalization could lead to an ``information cocoon'' or overfitting to a user's past behavior, affecting the diversity and fairness of the service. Additionally, achieving effective cold-start personalization without relying on a large amount of user data is also a challenge that needs to be addressed.

    \item \textbf{Scalability and Robustness in Real-World Settings:} Although the Alpha-Service system performs well in case studies, its robustness and scalability are still challenged in broader and more complex real-world scenarios, such as extreme lighting, noisy environments, and multi-user interactions. In addition, the stability of multi-module collaboration, error recovery mechanisms, and the ability to migrate services across different scenarios are all key issues that need to be addressed in future system iterations.

    \item \textbf{Privacy and Data Security:} The system continuously perceives the user's environment information through ego-centric videos, which inevitably involves the collection and processing of a large amount of personal privacy data. Even when we apply the localized storage, the long-term behavior recording and personalized preference learning may still raise concerns about privacy leakage. Ensuring that data is fully localization and anonymization while providing highly personalized services is key to gaining users' trust.
    
    \item \textbf{User Adaptation and Trust Building:} Users' acceptance and trust in proactive AI are key to the success of the system. Some users may feel uncomfortable with AI's active intervention or develop a dependency. The system should have an explainable decision-making mechanism in the future and allow users to provide feedback, correct inaccuracies, or disable services during interactions, thereby gradually building trust and promoting human-AI collaboration.

\end{itemize}

\section{Conclusion}
\label{sec:conclusion}
In this work, we introduced \textbf{\textit{AI for Service}} (AI4Service), a new paradigm for proactive AI assistance. We analyzed the limitations of existing reactive AI service systems and proposed the \textit{Alpha-Service} framework to address two central challenges: \textbf{\textit{Know When}} to act and \textbf{\textit{Know How}} to assist effectively. 
Drawing inspiration from the von Neumann computer architecture, our design integrates five key components that together offer a systematic foundation for building proactive assistants. 
We validated this concept through a multi-agent implementation on AI glasses, demonstrating its versatility across real-world scenarios such as real-time gaming assistance, museum guidance, and shopping support. 
These realistic case studies illustrate Alpha-Service's ability to perceive user intent, anticipate needs, and provide timely, meaningful assistance without explicit commands.
This work represents a step toward a more symbiotic form of human-AI interaction. 
In future research, we plan to enhance the personalization capacity of the Memory Unit, broaden the toolset of the Arithmetic Logic Unit, and conduct large-scale user studies to assess the long-term impact of proactive assistance in everyday contexts. Ultimately, we envision AI evolving into an indispensable and empathetic partner that truly understands and anticipates human needs.

\clearpage
\bibliographystyle{plainnat}
\bibliography{main}

\clearpage
\clearpage
\appendix

\section{Prompts for deciding whether using a web search or not.}
\label{appendix:prompts_decision}
Here we place the prompts of decision mechanism.
\begin{lstlisting}[style=promptstyle, caption={Decision Prompt}]
You are a helpful assistant. You can call tools. If you cannot answer my question or need help from the website, return the answer format of web_search(\"xxx\"). 
<question>
\end{lstlisting}

\section{Prompts for action instruction generation.}
\label{appendix:prompts_action_instruction}
Here we place the prompts of step I for Output Unit.
\begin{lstlisting}[style=promptstyle, caption={Action Instruction Generation Prompt}]
Here is a detailed analysis generated by the reasoning model. Please summarize it into a clear and concise action recommendation for the user.
Analysis Content:
<content>

Answer with one direct sentence:
\end{lstlisting}

\end{document}